%% file: acl2023.tex
\pdfoutput=1

\documentclass[11pt]{article}

\usepackage{ACL2023}

\usepackage{times}
\usepackage{latexsym}

\usepackage[T1]{fontenc}

\usepackage[utf8]{inputenc}

\usepackage{microtype}

\usepackage{inconsolata}

\usepackage{microtype}
\usepackage{microtype}
\usepackage{graphicx}
\usepackage{amsmath}
\usepackage{booktabs}
\usepackage{url} 
\usepackage{xcolor}
\usepackage{latexsym}
\usepackage{amsfonts}
\usepackage{amsmath}
\usepackage{bbm} 
\usepackage{amssymb}
\usepackage{subfigure}
\usepackage{balance}
\usepackage{threeparttable}
\usepackage{multirow}
\usepackage{float}
\usepackage{longtable}
\usepackage{lipsum}
\usepackage[noend]{algpseudocode}
\usepackage{algorithmicx,algorithm}
\usepackage{makecell}
\usepackage{bm}

\title{Reasoning over Hierarchical Question Decomposition Tree \\for Explainable Question Answering}

\author{
 Jiajie~Zhang$^{1}$\thanks{\ \ Indicates equal contribution.}\hspace{0.3em}, Shulin~Cao$^{2*}$, Tingjian~Zhang$^{2}$, Xin~Lv$^{2}$, Jiaxin~Shi$^3$, Qi~Tian$^3$, \\
 \textbf{Juanzi~Li$^2$, Lei~Hou$^2$}\\ 
 $^1$Institute for Interdisciplinary Information Sciences, Tsinghua University, Beijing, China \\
 $^2$Department of Computer Science and Technology, Tsinghua University, Beijing, China \\
 $^3$Cloud BU, Huawei Technologies\\
\texttt{\{jiajie-z19,  caosl19\}@mails.tsinghua.edu.cn}\\
\texttt{\{houlei, lijuanzi\}@tsinghua.edu.cn}\\
}

\begin{document}
\maketitle

\begin{abstract}
Explainable question answering (XQA) aims to answer a given question and provide an explanation why the answer is selected. Existing XQA methods focus on reasoning on a single knowledge source, e.g., structured knowledge bases, unstructured corpora, etc. However, integrating information from heterogeneous knowledge sources is essential to answer complex questions. In this paper, we propose to leverage question decomposing for heterogeneous knowledge integration, by breaking down a complex question into simpler ones, and selecting the appropriate knowledge source for each sub-question. To facilitate reasoning, we propose a novel two-stage XQA framework, \textit{Reasoning over Hierarchical Question Decomposition Tree} (RoHT). First, we build the \textit{Hierarchical Question Decomposition Tree} (HQDT) to understand the semantics of a complex question; then, we conduct probabilistic reasoning over HQDT from root to leaves recursively, to aggregate heterogeneous knowledge at different tree levels and search for a best solution considering the decomposing and answering probabilities.
The experiments on complex QA datasets KQA Pro and Musique show that our framework outperforms SOTA methods significantly, demonstrating the effectiveness of leveraging question decomposing for knowledge integration and our RoHT framework.
\end{abstract}

\input{1introduction.tex}

\input{2relatedwork.tex}

\input{3method.tex}

\input{4experiments.tex}

\input{5conclusion.tex}

\bibliography{anthology, custom}
\bibliographystyle{acl_natbib}

\clearpage

\appendix

\input{appendix.tex}

\end{document}

%% file: 1introduction.tex
\section{Introduction}
Explainable question answering (XQA) is the
task of (i) answering a question and (ii) providing
an explanation that enables the user to understand why the answer is selected~\cite{neches1985explainable,schuff2020f1}. It provides a qualified way to test the reasoning ability and interpretability of intelligent systems, and plays an important role in artificial intelligence~\cite{lulearn}.  

Recent work in XQA can be grouped into two directions: 1) neuro-symbolic methods~\cite{berant2013semantic,NSM,cao2022program} translate natural language questions into formal representations (\textit{e.g.}, SPARQL~\cite{sun2020sparqa}, KoPL~
\cite{cao2022kqa}, lambda-DCS~\cite{liang2013lambda}, \textit{etc.}), whose execution on structured knowledge bases (KBs) gives the answer. Here, the formal representation acts as an explanation of the final answer. 2) Decompose-based models generate natural language intermediate steps that lead to the final answer (e.g., question decomposing which decomposes a complex question into sub-questions~\cite{min2019multi,perez2020unsupervised,deng2022interpretable}, chain-of-thought prompting~\cite{Wei2022ChainOT,Dua2022SuccessivePF,Khot2022DecomposedPA}, etc.). Here, the intermediate steps shows the rationale of reasoning.

\begin{figure}[!t]
    \centering
    \includegraphics[width=0.48\textwidth]{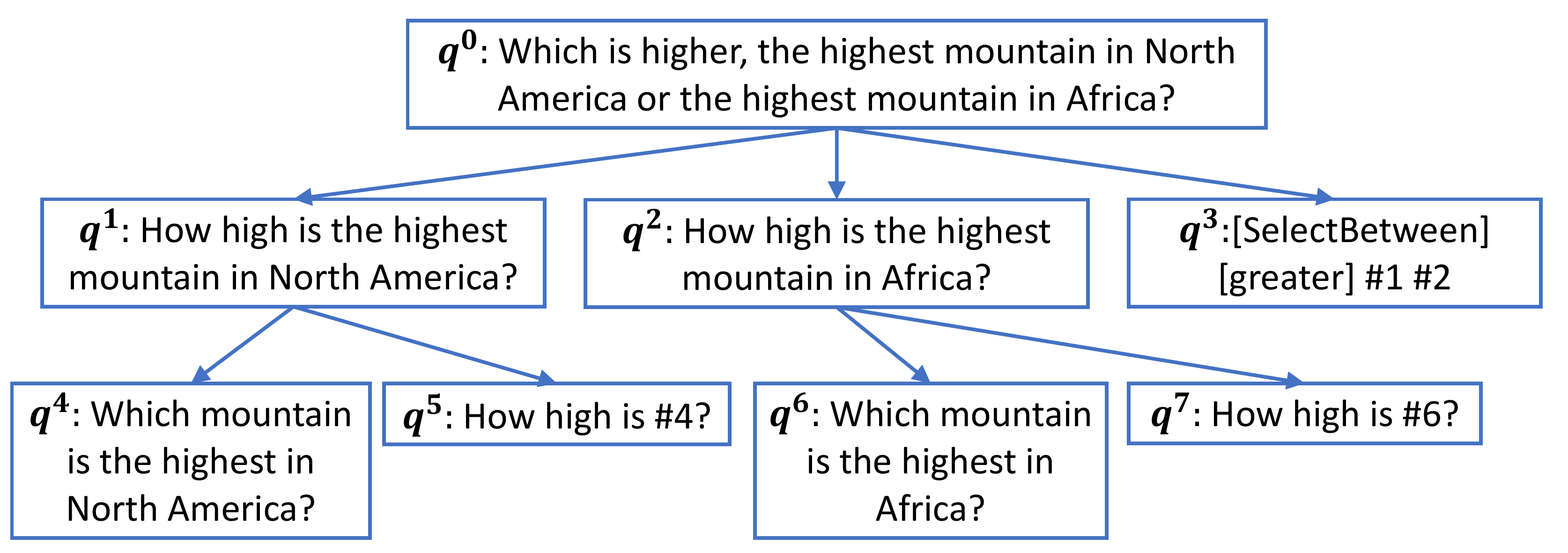}
    \caption{An example of Hierarchical Question Decomposition Tree (HQDT). $q^i$ represents the index of node in its BFS ordering enumeration.}
    \label{fig:hqdt}
\end{figure}

Although achieving significant results, both directions have key limitations. For neuro-symbolic methods, the formal representation can only be executed on KBs. However, even the largest KBs are incomplete, thus limits the recall of model. For decompose-based methods, they employ free-text corpora as the knowledge source, and the diversity of natural language makes XQA difficult.
In fact, integrating knowledge from
heterogeneous sources is of great importance to QA~\cite{wolfson2020break}, especially for answering complex questions. Several attempts have been made for knowledge integration (e.g., KBs, text corpora)~\cite{sun2018open,sun2019pullnet,shi2021transfernet}. Although promising, these graph-based methods suffer from lacking explainability or are constrained to limited reasoning capability.

Intuitively, leveraging question decomposing to integrate heterogeneous knowledge sources is a promising direction, since we can flexibly select the appropriate knowledge source for each sub-question. The challenges lie in: 1) How to determine the granularity of question decomposing, since certain complex questions can be directly answered with a knowledge source, and further decomposition increases the possibility of error. For example, in Figure~\ref{fig:hqdt}, $q^1$ can be answered with the Wikipedia corpus without further decomposition. 2) How to find the optimal solution among various possible ones, since question decomposing and answering are both uncertain. For example, $q^0$ can also be decomposed as ``Which mountains are in North America or Afirica'', ``\emph{What's the height of \#1}'', ``\emph{[SelectAmong] [largest] \#2}''.

To this end, we propose a novel two-stage XQA framework \textit{Reasoning over Hierarchical Question Decomspotion Tree}, dubbed RoHT. First, we propose to understand the complex question by building its \textit{hierarchical question decomposition tree} (\textbf{HQDT}). In this tree, the root node is the original complex question, and each non-root node is a sub-question of its parent. The leaf nodes are atomic questions that cannot be further decomposed. Compared with existing representations that directly decompose a question into the atomic ones, e.g., QDMR~\cite{wolfson2020break}, our tree structure provides the flexibility to determine solving a question whether by directly answering or further decomposing. Second, we propose \textbf{probabilistic reasoning over HQDT}, to fuse the knowledge from KB and text at different levels of the tree, and take into consideration the probability score of both tree generation and answering. The reasoning process is recursive, from the root to leaves, and constitues three steps: 1) a scheduler determines the appropriate knowledge sources for a particular question (from KB, text, or solving its children sequentially); 2) the corresponding executors output the answers with probabilities; 3) an aggregator aggregates the candidate answers from all the knowledge sources and outputs the best ones.

In evaluation, we instantiate our RoHT framework on two complex QA datasets: KQA Pro~\cite{cao2022kqa}, where we remove half of the triples in its KB and supplement it with Wikipedia corpus, and Musique~\cite{trivedi2021musique}, where we take Wikidata~\cite{vrandevcic2014wikidata} as additional KB besides the given text paragraphs. Experimental results show that, RoHT improves the performance significantly under the KB+Text setting, by 29.7\% and 45.8\% EM score on KQA Pro and Musique compared with existing SOTA model. In addition, compared with the decompose-based methods, RoHT improves the SOTA by 11.3\% F1 score on Musique.

\textbf{Our contributions} include: 1) proposing to leverage question decomposing to integrate heterogeneous knowledge sources for the first time; 2) designing a novel two-stage XQA famework RoHT by first building HQDT and then reasoning over HQDT; 3) demonstrating the effectiveness of our RoHT framework
through extensive experiments and careful ablation
studies on two benchmark datasets.

%% file: 2relatedwork.tex
\section{Related Work}
\subsection{QA over Text and KB}
Over time, the QA task has evolved into two main streams: 1) QA over unstructured data (e.g., free-text corpora like Wikipedia); 2) QA over structured data (e.g., large structured KBs like DBpedia~\cite{dbpedia}, Wikidata~\cite{vrandevcic2014wikidata}). As structured and unstructured data are intuitively complementary information sources~\cite{Ouz2020UniKQAUR}, several attempts have been
made to combines the best of both worlds. 

An early approach IBM Watson~\cite{Ferrucci2012IntroductionT} combines multiple expert systems and re-ranks them to produce the answer. ~\cite{Xu2016HybridQA} maps relational phrases to KB and
text simultaneously, and use an integer linear program model to provide a globally optimal solution. Universal schema based method~\cite{das2017question} reasons over both KBs and text by aligning them in a common embedded space. 
GraftNet~\cite{sun2018open} and its successor PullNet~\cite{sun2019pullnet} incorporate free text into graph nodes to make texts amenable to KBQA methods. TransferNet~\cite{shi2021transfernet} proposes the relation graph to model the label-form relation from KBs and text-form relation from corpora uniformly.

Although achieving promising results, these methods lack interpretability or are constrained to limited question type, \textit{i.e.}, TransferNet shows interpretability with transparent step transfering, however, it can only answer multi-hop questions, and cannot deal with questions that require attribute comparison or value verification. In contrast, our proposed framework shows great interpretability with HQDT and cover more question types.

\subsection{Question Decomposing}
\label{sec:qd}
For datasets, KQA Pro~\cite{cao2022kqa} proposes to decompose a complex question into a multi-step program KoPL, which can be executed on KBs. BREAK~\cite{wolfson2020break} proposes to decompose questions into QDMR, which constitutes the ordered list of steps, expressed through natural language. Musique~\cite{trivedi2021musique} is a QA dataset constructed by composing single-hop questions obtained from existing datasets, and thus naturally provides question decompositions.

For models, several attempts have been made for learning to decompose with weak-supervision, such as span prediction based method~\cite{min2019multi}, unsupervised sequence transduction method ONUS~\cite{perez2020unsupervised}, AMR-based method QDAMR~\cite{deng2022interpretable}. Another line of work is to employ large language models with in-context learning, such as Least-to-most Prompting~\cite{Zhou2022LeasttoMostPE}, decomposed prompting~\cite{Khot2022DecomposedPA}, successive prompting~\cite{Dua2022SuccessivePF}.

Compared with existing works, we are the first to design a hierarchical question decomposition tree for integrating information from multiple knowledge sources.

%% file: 3method.tex
\section{Definition of HQDT}


Formally, given a complex question,  its HQDT is a tree $T$. Each node $q^i \in T$ represents a question. For root node, it represents the given complex question, and for non-root nodes, it represents a sub-question of its parent node. The leaf nodes are simple ("atomic") questions that cannot be decomposed. Note that HQDT is a 3-ary ordered tree. As shown in Figure~\ref{fig:hqdt}, we enumerate the nodes of $T$ with BFS ordering, and $q^0$ is the root question. 

A question $q^i = \left\langle w_1,\cdots,w_j,\cdots, w_{|q^i|}\right\rangle$ can be categorized into one of the three types according to the token vocabulary:
1) natural language question (e.g., $q^4$: ``\emph{Which mountain is the highest in North America?}''), here, $w_{j} \in \mathcal{V}$, and $\mathcal{V}$ is the word vocabulary; 2) bridge question (e.g., $q^5$: ``\emph{How high is $\#4$?}''), here, $w_{j} \in \mathcal{V}\cup\mathcal{R}$, and $\mathcal{R}$ is the reference token vocabulary. In this question, ``$\#4$'' refers to the answer of $q^4$, which is the sibling question of $q^5$;
3) symbolic operation question (e.g., $q^3$: ``\emph{[SelectBetween][greater] $\#1$ $\#2$}''), here, $w_{j} \in \mathcal{V}\cup\mathcal{R}\cup\mathcal{O}$, and $\mathcal{O}$ is the vocabulary of pre-defined symbolic operations, which are designed for supporting various reasoning capacity (e.g., attribute comparison and set operation) and are shown in appendix \ref{app:op} in details. Note that all the bridge questions and symbolic operation questions are atomic questions and can only appear in leaf nodes.

For every non-leaf question $q^i$, we define two ordered lists: 
\begin{itemize}
    \item $q^i.children = \left\langle q^{st_i}, \cdots, q^{ed_i}\right\rangle$, which are children of $q^{i}$, successively indexed from $st_i$ to $ed_i$. For example, for question $q^1$ in Figure~\ref{fig:hqdt}, $q^{1}.children$ is $\left\langle q^4, q^5\right\rangle$. 
    \item $q^i.atoms = \left\langle a_{1}^i, \cdots, a_{n_i}^i\right\rangle$, which is a list of atomic questions deduced from the $n_i$ leaf nodes of the sub-tree rooted by $q^i$, by rearranging the reference tokens. For example, for $q^0$ in Figure~\ref{fig:hqdt}, its leaf nodes is $\left\langle q^4, q^5, q^6, q^7, q^3\right\rangle$, and correspondingly, $q^{0}.atoms$ is $\left\langle q^4, \tilde{q}^5, q^6, \tilde{q}^7, \tilde{q}^3\right\rangle$, with $\tilde{q}^5$ as ``\emph{How high is \#1?}'', $\tilde{q}^7$ as ``\emph{How high is \#3}'', and $\tilde{q}^3$ as ``\emph{[SelectBetween][greater] \#2 \#4}''. The detailed deduction algorithm is in appendix~\ref{app:ar} due to space limit. We also call $q^i.atoms$ the \textit{atomic representation} of $q^i$.
\end{itemize}
Specially, among $q^i.children$, $q^{st_i}, \dots, q^{ed_i-1}$ are all natural language questions, and $q^{ed_i}$ is either a bridge question or a symbolic operation question.
Answering $q^i$ is semantically equivalent to answering sub-questions in $q^i.children$ or in $q^i.atoms$ sequentially. The last question in $q^i.children$ or $q^i.atoms$ returns the answer of $q^i$.

\section{Methodology}
Our framework RoHT is composed of two stages:
1) Building HQDT. We understand the hierarchical compositional structure of a complex question $q^0$ by generating its HQDT $T$ with probability, where each question $q^i\in T$ has a score $p_g^i$ that represents the certainty of its generation. 

2) Probabilistic Reasoning over HQDT. We conduct recursive probabilistic reasoning over the HQDT from root to leaves to solve $q^0$. For each question $q^i$, we will utilize KBs, text and its child questions together to get a list $R^i$, which contains answers of $q^i$ with probabilistic scores. Finally the answer with the highest score in $R^0$ will be picked out as the final answer of $q^0$.

The details are introduced as follows.


\subsection{Building HQDT}

\begin{figure*}[!t]
    \centering
    \includegraphics[width=\textwidth]{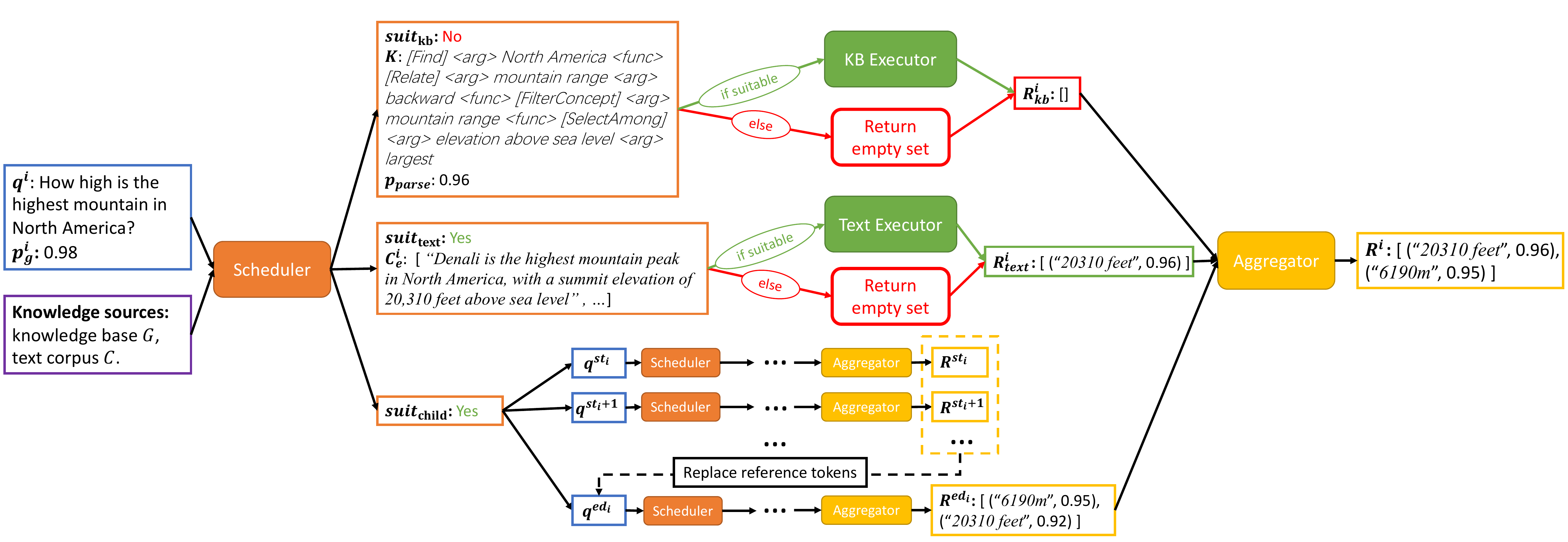}
    \caption{Illustration of the recursive reasoning function $f$. For a question $q^i$, $f$ uses the scheduler to determine suitable knowledge sources and calls executors to retrieve answers from them. $f$ also recursively calls itself to get answers from the children of $q^i$. Finally the answers from different sources are fused by the aggregator. }
    \label{fig:reason}
\end{figure*}
To build the HQDT for a complex question, we first generate its atomic representation, which corresponds the leaf nodes of HQDT, then generate every non-leaf nodes based on this atomic representation. We compute  certainty score of each node based on the likelihood of each step of generation. 

\paragraph{Building Leaf Nodes}
Given a complex question $q^0$, we first use a BART~\cite{lewis2019bart}-based question decomposer $M_\theta$ to generate its atomic representation and output the likelihood of generation:


\begin{equation}
    L^0, l_d = M_\theta(q^0).
\end{equation}
Here, $L^0=a_1^0\ \langle sep \rangle\ a_2^0\ \langle sep \rangle \dots \langle sep \rangle\ a_{n_0}^0$ is the serialization of $q^0.atoms$, where $\langle sep \rangle$ is a separating token. $l_d=\operatorname{Pr}(L^0|q^0; \theta)$ is the likelihood of generation.
Since $q^0$ is the root of $T$, each atomic question in $q^0.atoms$ corresponds to a leaf node in $T$ (with the deterministic algorithm in Appendix~\ref{app:generation}), and the certainty score of each leaf node in $T$ is $l_d$.


\paragraph{Building Non-leaf Nodes}
Based on $q^0.atoms$, we can generate all the non-leaf questions in HQDT. The root question is just $q^0$ and thus has certainty score $p_g^0=1$. For every other non-leaf question $q^i$, its atomic representation $q^i.atoms=\langle a^i_1,\dots, a^i_{n_i}\rangle$ can be translated from a specific subset of $q^0.atoms$ by rearranging the reference tokens. 
The subset can be determined by considering the reference relations of a bridge or symbolic operation question $a_j^0\in q^0.atoms$, which corresponds to the leaf node $q^{ed_i}$, with other questions in $q^0.atoms$. We show the details in Appendix~\ref{app:generation}. 
For example, $q^2.atoms$ in Figure~\ref{fig:hqdt} is (``\emph{Which mountain is the highest in Africa?}'', ``\emph{How high is $\#1$?}''), and it can be obtained from $(a_3^0, a_4^0)$ in $q^0.atoms$. 
Then we can use a BART-based question generator $M_\phi$ to generate $q^i$ from $q^i.atoms$:
\begin{equation}
    q^i, l_g^i = M_\phi(L^i), 
\end{equation} 
where $L^i=a_1^i\ \langle sep \rangle\ a_2^i\ \langle sep \rangle \dots \langle sep \rangle\ a_{n_i}^i$ is the serialized $q^i.atoms$, and $l_g^i=\operatorname{Pr}(q^i|L^i; \phi)$ is the likelihood of $q^i$ given $L^i$. The certainty score of $q^i$ is computed as:
\begin{equation}    
    p^i_g=l_d \cdot l_g^i.
\label{eq:generate}
\end{equation}

\paragraph{Learning of Question Decomposer and Generator}
The question decomposer $M_\theta$ can be trained with paired ($q^0$, $q^0.atoms$) data, where the atomic representation can be from either given annotation or unsupervised construction. The question generator $M_\phi$ can also be trained with the same data by exchanging the input and output. The details are shown in Section \ref{sec:implement}.

\subsection{Probabilistic Reasoning over HQDT}


\begin{equation}
        f(q^i, p_g^i, G, C) \rightarrow R^i:\{(ans_j^i, p_j^i)\},
\end{equation}
where $ans_j^i$ is an answer of $q^i$, and score $p_j^i$ represents the certainty of $ans_j^i$. 

As shown in Figure~\ref{fig:hierarchical}, the implementation of $f$ contains tree steps: 1) a scheduler determines the suitable knowledge sources for a particular question, i.e., whether the question can be answered from KB, text, or by solving its child questions sequentially; 2) according to the suitable sources output by the scheduler, executors aim to get the answers with probabilities via executing on KB (KB executor) or retrieving from text (text executor), or answering the child questions (call $f$ recursively); 3) an aggregator aggregates candidate answers from all the knowledge sources and outputs the $\operatorname{top}$-$k$ answers according to their probabilities. In the following, we will introduce their details when answering $q^i$.

\paragraph{Scheduler} We formalize the scheduler as:
\begin{equation}
    \begin{aligned}
        &suit_\text{kb}, suit_\text{text}, suit_\text{child} =\operatorname{Scheduler}(q^i, G, C),
    \end{aligned}
\end{equation}
Where $suit_\text{kb}$, $suit_\text{text}$ and $suit_\text{child}$ are 0/1 variables, respectively representing whether the answers of $q^i$ are suitable to get from the KB $G$, the corpus $C$, or by solving $q^i.children$ sequentially.

Specifically, to check whether $G$ is suitable, the scheduler employs a semantic parser~\cite{cao2022kqa} $M_\text{sp}$ to parse $q^i$ into a program $K$ with probability $p_\text{parse}$:
\begin{equation}\label{eq:kb}
    K, p_\text{parse} = M_\text{sp}(q^i).
\end{equation}
Then it classifies the type of $q^i$ according to the function skeleton of $K$. For example, the function skeleton of $K$ in Figure~\ref{fig:reason} is ``\emph{Find-Relate-FilterConcept-SelectAmong}''. If the precision of $G$ on the questions that have the same function skeleton with $K$ is larger than a predefined threshold $\gamma$ \footnote{The precision of KB is calculated with questions in training set}, the scheduler will set $suit_\text{kb}$ to be 1. 

To check whether the corpus $C$ is suitable, the scheduler tries to find a set of evidence paragraphs for $q^i$. If $C$ is too large, the scheduler will first use BM25~\cite{robertson2009probabilistic} to recall dozens of most relevant paragraphs. For each paragraphs, we train a RoBERTa~\cite{liu2019roberta}-based selector $M_\text{sl}$ to classify whether it is an evidence paragraph for $q^i$. Suppose the set of selected evidence paragraphs, $C_{e}$ is not empty, the scheduler will set $suit_\text{text}$ as 1. 

To make best use of knowledge from all levels, the scheduler simply set $suit_\text{child}$ to be 1 if $q^i$ is a non-leaf question otherwise 0.

\paragraph{Executors}

For the KB executor, it takes the program $K$ in Equation~\ref{eq:kb} on KB $G$ to get the answers, and takes the parsing score $p_\text{parse}$ in Equation~\ref{eq:kb} to calculate the probability score for each answer:
\begin{equation}
    \begin{aligned}
        &R_\text{kb}^i = \{(ans_{\text{kb}, j}^i, \ p^i_g\cdot p_\text{parse})\}.
    \end{aligned}
\end{equation}

For the text executor, it takes the selected paragraph set $C_e$ as described above, and
employs a Transformer-based reading comprehension model $M_\text{rc}$ to extract answers from $C_e$:
\begin{equation}
    \begin{aligned}
        & \{(ans_{{\text{text}, j}}^i, p_{\text{ex}, j}^i)\}=M_\text{rc}(q^i, C_e), \\
        & R_\text{text}^i = \{(ans_{\text{text}, j}^i, \ p^i_g\cdot p_{\text{ex}, j}^i)\}.
    \end{aligned}
\end{equation}
where $p_{\text{ex}, j}^i$ is the extraction probability of $ans_{\text{text}, j}^i$ given by $M_\text{rc}$. 

For solving $q^i$ by answering its children, $f$ will recursively call itself to solve $q^{st_i}, \dots, q^{ed_i}$ in order:
\begin{align}
    &R^{st_i} = f(q^{st_i}, p_g^{st_i}, G, C), \nonumber\\
    &R^{st_i+1} = f(q^{st_i+1}, p_g^{st_i+1}, G, C),  \\
    &\ \ \ \ \ \ \  \dots \nonumber\\
    &R^{ed_i} = f_\text{ref}(q^{ed_i}, p_g^{ed_i}, G, C, [R^{st_i}, \dots, R^{ed_i-1}]),\nonumber 
\end{align}
and let 
\begin{equation}
    R_\text{child}^i=R^{ed_i}.
\end{equation}
Here, $f_\text{ref}$ is a variant of $f$ to solve bridge and symbolic questions, which refer to the answers of their sibling questions.
Suppose $q^{ed_i}$ refers to the answers of its siblings $q^{r_1}, \dots, q^{r_{h_i}}$ in order. If $q^{ed_i}$ is a bridge question, $f_{ref}$ will 1) convert $q^{ed_i}$ into several possible natural language question $q_\text{nl}^1, \dots, q_\text{nl}^K$ by replacing the reference tokens with every combination $((x_1^k, v_1^k), \dots, (x_{h_i}^k, v_{h_i}^k)) \in R^{r_1}\times \dots \times R^{r_{h_i}}$, 2) call $f$ to solve each $q_\text{nl}^k$ and 3) fuse the answers from each $R_\text{nl}^k$ and select the top-$k$ answers with the highest scores:
\begin{align}
    &\{(ans_{\text{nl}, j}^k,\ p_{\text{nl}, j}^k)\} = f(q_\text{nl}^j, p_g^i, G, C), \nonumber\\
    &R_\text{nl}^k = \{(ans_{\text{nl}, j}^k,\ \operatorname{Avg}(p_{\text{nl}, j}^k, v_1^k, \dots, v_{h_i}^k))\}, \nonumber\\
    & R^{ed_i} = \operatorname{Select}(R_\text{nl}^1, \dots, R_\text{nl}^K)
\end{align}
Note that the score of answer $ans_{\text{nl}, j}^k$ is computed by averaging $p_{\text{nl}, j}^k$ and $v_1^k, \dots, v_{h_i}^k$, instead of multiplying them, to avoid exponential shrink during recursion. If $q^{ed_i}$ is a symbolic operation question with operation $op$ and arguments, $f_\text{ref}$ will execute simple program to apply the operation $op$ over $R^{r_1}, \dots, R^{r_{h_i}}$ to get $R^{ed_i}$. The score of each answer $ans_j^{ed_i}$ is computed as the average of $p_g^{ed_i}$ and the scores of answers in $R^{r_1}, \dots, R^{r_{h_i}}$ used by the program to get $ans_j^{ed_i}$.

\paragraph{Aggregator}
The aggregator fuses $R_\text{kb}^i$, $R_\text{text}^i$ and $R_\text{child}^i$ by selecting the $\operatorname{top}$-$k$ answers with the highest scores from them. 
If several answers have the same surface form, only the one with the highest score will be preserved.   

\begin{align}
    & R^{i} = \operatorname{Aggregator}(R_\text{kb}^i, R_\text{text}^i, R_\text{child}^i).
\end{align}

%% file: 4experiments.tex
\section{Experiments}
\subsection{Datasets}
Currently, there are few high-quality complex QA datasets based on both KBs and text. Previous methods~\cite{sun2018open, sun2019pullnet, shi2021transfernet} evaluated their models on MetaQA~\cite{zhang2018variational} by pairing its KB with the text corpus of WikiMovies~\cite{miller2016key}. However, the questions in MetaQA are too simple since there are only 9 relations in its KB. Therefore, we conduct our experiments on two more challenging complex QA datasets: KQA Pro and Musique, and their details are as follows. 

\textbf{KQA Pro}~\cite{cao2022kqa} is a large scale complex QA dataset, including 120k diverse natural language questions up to 5 hops over KB. 
Its KB is a subset of Wikidata~\cite{vrandevcic2014wikidata}, and consists of 16k entities, 363 predicates, 794 concepts and 890k triple facts. For each question, KQA Pro also provides the corresponding KoPL program. To simulate the realistic case where KB is incomplete, following~\cite{sun2019pullnet, shi2021transfernet}, we randomly discard 50\% triples in the KB and take Wikipedia as supplementary text corpus. 

\textbf{Musique}~\cite{trivedi2021musique} is a multi-hop QA dataset over text, including 25k 2-4 hop questions. We evaluate our framework under Musique-Ans setting where all the questions are answerable. Its questions are carefully constructed from several single-hop QA datasets via manually composition and paraphrase, and are hard to cheat via reasoning shortcut. For each complex question, Musique gives 20 paragraphs (including annotated evidence paragraphs and distractor paragraphs) as the corpus. Specially, for each question in the training set, Musique also provides a golden atomic representation, together with the answer and the evidence paragraph of each atomic question. In addition to the given paragraphs, we choose Wikidata as the KB to acquire additional knowledge.

\begin{table*}[!t]
\centering
\begin{tabular}{lcccccccc}
\toprule
Model         & \textbf{Overall} & Multihop       & Qualifier      & Comparison     & Logical        & Count & Verify         & Zero-shot      \\\hline
\multicolumn{9}{l}{\textit{50\% KB}}                                                                                                           \\\hline
KVMemNN       & 17.72            & 17.63          & 18.53          & 1.39           & 15.48          & 28.38 & 59.30          & 0.06           \\
RGCN          & 34.77            & 33.71          & 28.44          & 31.46          & 35.39          & \textbf{39.76} & 64.27          & 0.06           \\
BART KoPL     & 38.04            & 33.10          & 29.40           & 51.81          & 29.92          & 33.69 & 60.12          & 29.03          \\
$\text{RoHT}^\text{KB}$ & 38.94         & 34.16          & 31.54          & 50.91     &31.61     & 33.69 & 60.4        & 30.52 \\\hline
\multicolumn{9}{l}{\textit{50\%KB + Text}}                                                                                                     \\\hline
TransferNet   & 16.80            & 15.94          & 17.93          & 45.35 & 14.84          & 10.47 & 0.00           & 8.43           \\
$\text{RoHT}^\text{mix}$ & \textbf{46.45}   & \textbf{41.76} & \textbf{41.73} & \textbf{52.21}          & \textbf{41.95} & 31.26 & \textbf{65.45} & \textbf{38.76} \\\bottomrule
\end{tabular}
\caption{EM results on the dev set of KQA Pro. RoHT outperforms all the baselines by a large margin and achieves the best performance on most types of questions. }
\label{table:KQA Pro}
\end{table*}

\subsection{Implementations}
\label{sec:implement}
\paragraph{KQA Pro}
For the experiments of KQA Pro, a key challenge is that there are no annotations for atomic representation, which are required for training the question decomposer and generator in RoHT. Because the KoPL program of a complex question follows context free grammar, every atomic question will correspond to a specific span of the program. Therefore we first split the KoPL program into sub-programs according to the grammar, then use each sub-program to generate the atomic question by applying BART model fintuned with the (KoPL, question) pairs from the original dataset. For the answers for each atomic question, we execute the corresponding sub-programs on the KB to get corresponding answers. 
Using these constructed atomic representations, we train two BART-base models as the question decomposer and generator, respectively. 

For the scheduler, we directly use the semantic parser trained by~\cite{cao2022kqa} on KQAPro, and set the precision threshold $\gamma$ to be 0.7. We train a RoBERTa-large as the evidence selector via weak supervised method: for each question in the training set and constructed atomic representations, we first use BM25 to recall 10 related paragraphs from wikipedia, then take the paragraphs that contain the answer as positive samples and take other recalled paragraphs as negative samples.
For the text executor, we also train a BART-large reading comprehension model on these positive samples.

\paragraph{Musique}
Since Musique provides golden atomic representation for every complex question in the training set, we directly use them to train BART-base models as question decomposer and generator. For the scheduler, we adapt semantic parser trained by~\cite{cao2022kqa} on Wikidata. The KB precision threshold $\gamma$ is set to be 0.4, which is determined by the top-10 types of questions with the highest precision. We train the RoBERTa selector model on complex and atomic questions in the training set together, taking annotated evidence paragraphs as positive samples and distractor paragraphs as negative samples. For the text executor, we pre-train a Longformer-large~\cite{beltagy2020longformer} reading comprehension model on SQUAD~\cite{rajpurkar2016squad}, then finetune it on complex questions and atomic questions of Musique.

\subsection{Baselines}
we compare RoHT with several representative methods for complex QA, including memory-based methods, graph-based methods, and XQA methods.\\
\textbf{KVMemNN}~\cite{miller2016key} stores encoded knowledge in key-value memory and iteratively reads the memory to update the query vector to conduct multi-hop reasoning. \\
\textbf{RGCN}~\cite{schlichtkrull2018modeling} is a variant of graph convolutional network and utilizes the graph structure of KB to tackle complex questions. \\
\textbf{BART KoPL}~\cite{cao2022kqa} is a BART-based semantic parser which can convert complex question into KoPL program. It achieves over 90\% accuracy on KQA Pro on the complete KB. \\
\textbf{SA}~\cite{trivedi2021musique} is a two-stage model that first uses a RoBERTa-large selector to rank and select the $K$ most relevant paragraphs with the question and then uses a Longformer-large answerer to predict answer based on selected paragraphs. \\
\textbf{EX(SA)}~\cite{trivedi2021musique} is the state-of-the-art model on Musique. It first explicitly decomposes the complex question into atomic representation and then calling SA model repeatedly to answer each atomic question in order. \\
\textbf{TransferNet}~\cite{shi2021transfernet} iteratively transfer entity scores via activated path on the relation graph that consists of both text-form relations and KB-form relations. It is existing state-of-the-art model that utilizes both KBs and text as knowledge soruces, and nearly solves MetaQA. We reimplement it on both KQA Pro and Musique, and the details are shown in Appendix~\ref{app:transfernet}. \\
\textbf{RoHT}: $\text{RoHT}^\text{KB}$, $\text{RoHT}^\text{text}$ and $\text{RoHT}^\text{mix}$ denote the RoHT models that only use KB, only use text and use both KB and text, respectively.

\subsection{Main Results}

\subsubsection{Results on KQA Pro}
The experimental results for KQA Pro are shown in Table \ref{table:KQA Pro}. When using only the incomplete KB, $\text{RoHT}^\text{KB}$ model respectively improves EM by 21.22, 4.17 and 0.90 compared to KVMemNN, RGCN and BART KoPL, showing the benefit of integrating the answers of sub-questions of different levels. After adding Wikipedia as supplementary text corpus, $\text{RoHT}^\text{mix}$ yields substantial improvement compared with $\text{RoHT}^\text{KB}$ (7.51 on EM), demonstrating the effectiveness of utilizing knowledge from KB and text together. 
$\text{RoHT}^\text{mix}$ also outperforms TransferNet, which is end-to-endly trained with a mixed relation graph, by a large margin (29.65 on EM). This is because unlike graph-based methods, RoHT explicitly shows the compositional structure of a complex question in natural language form via HQDT generation, and thus can retrieve answers from the KB and text with more advanced and flexible sub-modules (e.g., semantic parser and reading comprehension model).  
Moreover, our designed atomic operations in the HQDT also enable RoHT to solve a wide variety of complex questions: we can see that $\text{RoHT}^\text{mix}$ achieves the best results on 6 types of questions among 7 types, showing comprehensive reasoning capacity.

\begin{table}[!t]
\centering
\begin{tabular}{lcc}
\toprule
Model           & EM       & F1      \\\hline
\multicolumn{3}{l}{\textit{Text}}    \\\hline
SA              & 39.3     & 47.3    \\
EX(SA)          & 41.5     & 49.7    \\
$\text{RoHT}^\text{text}$ & 53.1 & 61.6 \\\hline
\multicolumn{3}{l}{\textit{Text+KB}} \\\hline
TransferNet     & 8.6      & 10.9    \\
$\text{RoHT}^\text{mix}$            & \textbf{54.4}     & \textbf{62.3}   \\\bottomrule
\end{tabular}
\caption{EM and F1 results on the dev set of Musique. Compared with state-of-the-art methods, RoHT achieves significant improvement.}
\label{table:musique}
\end{table}

\subsubsection{Results on Musique}
Table \ref{table:musique} presents the results on the dev set of Musique dataset. As expected, our RoHT models show significant improvement over all the baselines. With only given paragraphs, $\text{RoHT}^\text{text}$ improves EM/F1 by 13.8/14.3 and 11.6/11.9 compared with SA and EX(SA), respectively; With both text and KB, the performance of $\text{RoHT}^\text{mix}$ is also remarkably better than TransferNet (62.3 v.s. 10.9 on F1). Comparing $\text{RoHT}^\text{text}$ and $\text{RoHT}^\text{mix}$, we can also see some benefits of supplementing the text information with KB information, though the improvement is smaller than supplementing the KB with text on KQA Pro because KBs have lower coverage than text and the semantic parser is not specially finetuned for questions of Musique. We submit the predictions of $\text{RoHT}^\text{mix}$ on the test set and achieve 63.6 F1 score, which significantly outperforms the best public result 52.3.

\subsection{Further Analysis}
\subsubsection{Effect of Scheduler}
\begin{table}[!t]
\centering
\begin{tabular}{l|c|c}
\hline
\textbf{Model}  & \textbf{KQA Pro} & \textbf{Musique} \\ \hline
$\text{RoHT}^\text{mix}$           & 46.5            & 54.4             \\ 
w/o scheduler  & 40.7            & 47.0                \\ 
$\text{RoAT}^\text{mix}$  & 32.3            & 47.6                \\ \hline
\end{tabular}
\caption{EM performance of $\text{RoHT}^\text{mix}$ with and without scheduler, and EM performance of $\text{RoAT}^\text{mix}$.}
\label{table:analysis}
\end{table}
To show the effect of the scheduler module, we remove it from the $\text{RoHT}^\text{mix}$ model, i.e, default that the KB and recalled/given text paragraphs are suitable for all questions in the HQDT, and evaluate the performance again on the dev set of KQA Pro and Musique. The results are shown in Table~\ref{table:analysis}. We can see that after discarding the scheduler, the EM performance on KQA Pro and Musique drops by 5.8 and 7.4, respectively. Therefore, it is important to use the scheduler to select suitable knowledge sources for each question.

\subsubsection{Effect of Hierarchical Decomposition}

\begin{figure}[!t]
    \centering
    \includegraphics[width=0.48\textwidth]{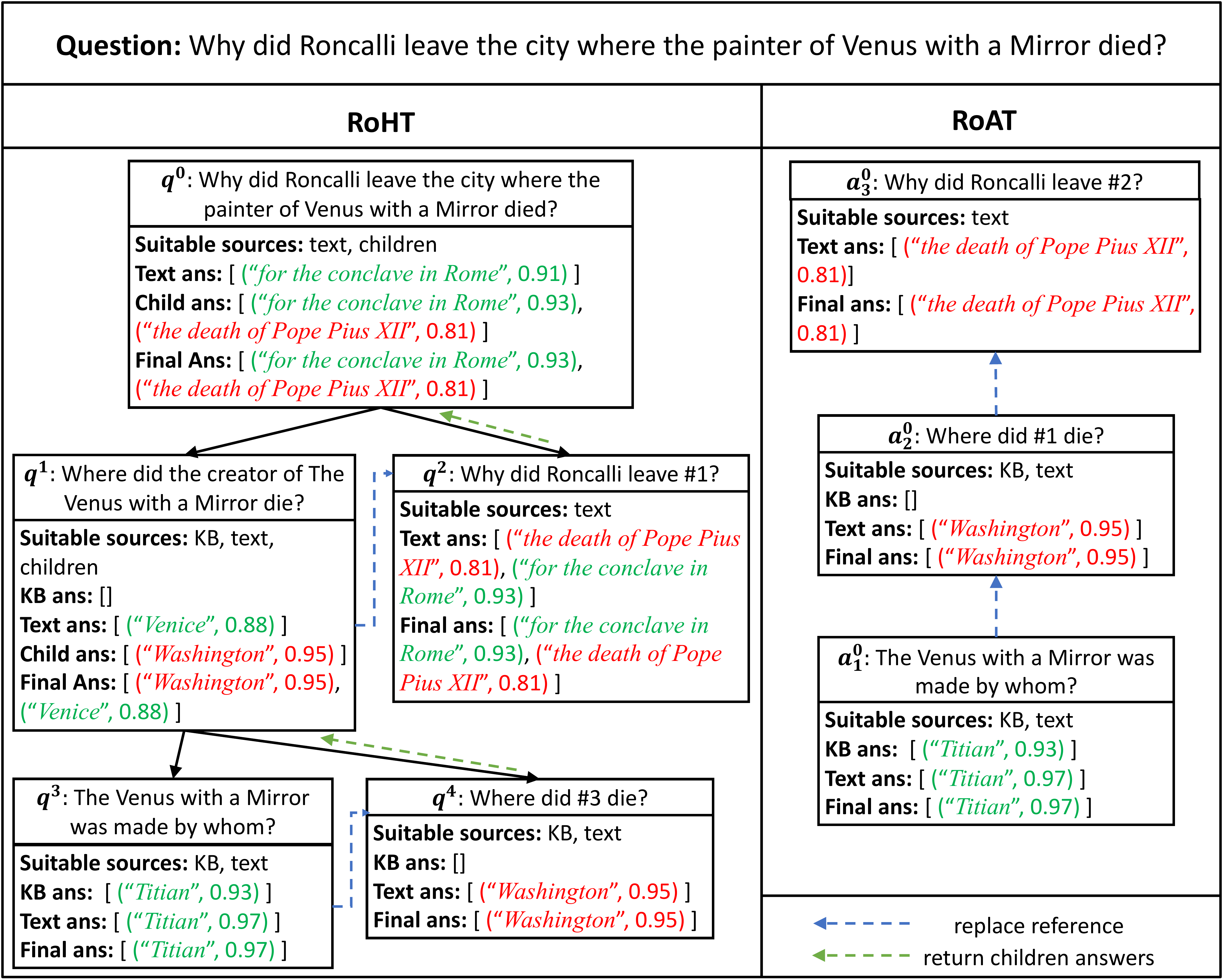}
    \caption{A case from Musique. We mark the correct answers in green and the wrong answers in red.}
    \label{fig:hierarchical}
\end{figure}
Many existing methods generate non-hierarchical decomposition of complex questions, similar to the atomic representation, to assist reasoning~\cite{min2019multi, wolfson2020break, deng2022interpretable}.  To demonstrate the superiority of hierarchical decomposition, we compare our $\text{RoHT}^\text{mix}$ model with $\text{RoAT}^\text{mix}$ model, which uses the same scheduler, executors, and aggregator as $\text{RoHT}^\text{mix}$, but solves the complex question by directly answering the atomic questions in its atomic representation in order. As shown in Table~\ref{table:analysis}, $\text{RoHT}^\text{mix}$ outperforms $\text{RoAT}^\text{mix}$ by a large margin on both KQA Pro and Musique. This is because the hierarchical structure of HQDT enables RoHT model to fuse the knowledge from KBs and text at different question levels, and to discard wrong answers via comparing the problisitic scores of answers. 

To further understand the reason, we show a case from Musique in Figure~\ref{fig:hierarchical}. We can see that both $\text{RoHT}^\text{mix}$ and $\text{RoAT}^\text{mix}$ fail to answer the question ``\emph{Where did (Titian) die?}'' ($q^4$ in the left, $a_2^0$ in the right). However, $\text{RoHT}^\text{mix}$ directly extracts the correct answer of $q^1$ from text and finally gets the correct answer of $q^0$ with the highest score,  while $\text{RoHT}^\text{mix}$ fails to solve $a_3^0$ because it must rely on the wrong answer from $a_2^0$.

%% file: 5conclusion.tex
\section{Conclusion}
In this paper, we propose RoHT, an understanding-reasoning XQA framework that uses both a KB and a text corpus to derive answers of complex questions. RoHT first builds the HQDT for a complex question to understand its hierarchical compositional structure, then conducts recursive probabilistic reasoning over the HQDT to solve the question, integrating answers from the KB, text, and sub-questions. Experiments show that RoHT significantly outperforms previous methods. We also demonstrate the superiority of HQDT compared with non-hierarchical decomposition.


\section{Limitation}
 Currently, RoHT framework is restricted to incorporating KBs and text. However, since RoHT retrieves answers from each knowledge source in a separate way, it could in principle utilize knowledge from more heterogeneous sources such as tables, and we will study this in future work. In addition, a device with large storage space and memory is needed for the storage and usage of Wikipeida and Wikidata.
 
\section{Ethics Statement}
The data used in this paper are drawn from publicly published datasets, encyclopedias and knowledge bases. Most of them do not involve sensitive data. 

%% file: appendix.tex
\section{Atomic Operations}
\label{app:op}

We design 6 atomic operations: \emph{Verify}, \emph{SelectBetween}, \emph{SelectAmong}, \emph{Count}, \emph{Intersection}, \emph{Union}, to support various reasoning capacity. We show their input, output, and examples in Table~\ref{table:operator}.

\begin{table*}
\scriptsize
\centering
\begin{tabular}{lccc}
\toprule
\textbf{Operation}     & \textbf{Argument}                        & \textbf{Input $\rightarrow$ Output}               & \textbf{Example}                                    \\ \hline
Verify                & Value, > / < / = / !=            & (Value) $\rightarrow$ (Bool)                      & \makecell[c]{{[}Verify{]} {[}2005{]} {[}<{}{]} \#3 \\ (1998) $\rightarrow$ (yes)}      \\ \hline
SelectBetween         & greater / smaller                        & (Value, Ent) (Value, Ent) $\rightarrow$ (ENT)     & \makecell[c]{{[}SelectBetween{]} {[}smaller{]} \#3 \#4 \\ (6670 km, Nile River) (6440 km, Amazon River) $\rightarrow$ (Amazon River)}         \\ \hline
SelectAmong           & largest / smallest                       & {[}(Value, Ent){]} $\rightarrow$ (Ent)            & \makecell[c]{{[}SelectAmong{]} {[}largest{]} \#1 \\ {[}(8848m, Everest) (8611m, K2) (8516m, Makalu){]} $\rightarrow$ (Everest)}                \\ \hline
Count                 & /                                        & {[}(Ent){]} $\rightarrow$ (Value)                 & \makecell[c]{{[}Count{]} \#2 \\ {[}(Bronny James) (Bryce James) (Zhuri James){]} $\rightarrow$ (3)   }                                 \\ \hline
Intersection          & /                                        & {[}(Ent){]} {[}(Ent){]} $\rightarrow$ {[}(Ent){]} & \makecell[c]{{[}Intersection{]} \#1 \#2 \\ {[}(apple) (orange) (peach){]} {[}(orange){]} $\rightarrow$ {[}(orange){]}   }                      \\ \hline
Union                 & /                                        & {[}(Ent){]} {[}(Ent){]} $\rightarrow$ {[}(Ent){]} & \makecell[c]{{[}Union{]} \#1 \#2 \\ {[}(apple) (orange){]} {[}(orange) (peach){]} $\rightarrow$ {[}(apple) (orange) (peach){]}   }                                 \\ \hline
\end{tabular}
\caption{Atomic operations proposed for QDKT, along with the corresponding examples. Ent, Value, Pred, and Bool denote entity, attribute, predicate, and boolean variable, respectively. () means tuple and [] means list.}
\label{table:operator}
\end{table*}

\section{Get Atomic Representation from Leaf Nodes}
\label{app:ar}
Algorithm~\ref{algo:atomic} describes that how to get the atomic representation of a question $q^i\in T$ from the leaf nodes in the sub-tree rooted by $q^i$.
\begin{algorithm}
\small
\caption{Get Atomic Representation from Leaf Nodes} 
\hspace*{0.02in} {\bf Input:} 
An HQDT $T$ and a index $i$. \\
\hspace*{0.02in} {\bf Output:} 
$q^i.atoms$\\
\begin{algorithmic}[1]
\Function{Dfs}{$j$, $atoms$, $ids$, $n$}
    \If{$q^j$ is a leaf question}
        \State $n\gets n+1$
        \State $ids[j]\gets n$
        \State $a\gets q^j$
        \For{$k$ in $\text{\textbf{GetRefTokens}}(q^j)$}
            \If{$q^k$ is a leaf question}
                \State $a\gets\text{\textbf{ModifyRefTokens}}(a, k, ids[k])$
            \Else
                \State $a\gets\text{\textbf{ModifyRefTokens}}(a, k, ids[ed_k])$
            \EndIf
        \EndFor
        \State $atoms.\text{append}(a)$
        \State \Return
    \EndIf
    \For{$k\gets st_j, \dots, ed_j$}
        \State $\text{\textbf{Dfs}}(k)$
    \EndFor
\EndFunction
\State 
\State $q^i.atoms \gets []$
\State $ids\gets$ empty dict
\State $\text{\textbf{Dfs}}(i, q^i.atoms, ids, 0)$
\end{algorithmic}
\label{algo:atomic}
\end{algorithm}

\section{Pseudocode for Building HQDT}
Algorithm~\ref{algo:build-tree} shows the pseudocode for generating the HQDT of a complex question with probability.
\label{app:generation}
\begin{algorithm}[!t]
\small
\caption{Generation of HQDT} 
\hspace*{0.02in} {\bf Input:} 
a complex question $q^0$, a question decomposer $M_\theta$, a question generator $M_\phi$. \\
\hspace*{0.02in} {\bf Output:} 
a list $T$ representing the HQDT, where element $(q^i, p_g^i, fa_i)$ in $T$ denote a sub-question $q^i$, certainty score of $q^i$ and the father of $q^i$, respectively.\\
\begin{algorithmic}[1]
\Function{RearrangeRefTokens}{$ar$}
    \State $atoms\gets []$
    \State $ids\gets$ empty dict
    \State $h\gets 0$
    \For{$(i, a_i)$ in $ar$}
        \State $h\gets h+1$
        \State $ids[i]\gets h$
        \For{$k$ in $\text{\textbf{GetRefTokens}}(a_i)$}
            \State $a_i\gets\text{\textbf{ModifyRefTokens}}(a_i, k, ids[k])$
        \EndFor
        \State $atoms.\text{append}(a_i)$
    \EndFor
    \State \Return $atoms$
\EndFunction
\State 
\State $([a_1^0, \dots, a_{n_0}^0], l_d)\gets M_\theta(q^0)$
\State $n\gets n_0$
\State $T \gets []$
\For{$i \gets 1,2,\dots, n_0$}
    \State $(q^i, p_g^i)\gets (a_i^0, l_d)$
    \State $ar^i \gets [(i, a_i^0)]$
    \If{$a_i^0$ contains referring tokens}
        \State $r_1, \dots, r_h \gets \text{\textbf{GetRefTokens}}(a_i^0)$
        \State $n\gets n+1$
        \State $ ar^n\gets []$
        \For{$j \gets r_1, \dots, r_h, i$}
            \If{$fa_j$ has been identified}
                \State $q^i\gets\text{\textbf{ModifyRefTokens}}(q^i, j, fa_j)$
                \State $j\gets fa_j$
            \EndIf
            \State $fa_j \gets n$
            \State $T\text{.append}((q^j, p_g^j, fa_j))$
            \State $ar^n\text{.extend}(ar^j)$
        \EndFor
        \State $q^n.atoms\gets \text{\textbf{RearrangeRefTokens}}(ar^n)$
        \State $L^n=\text{\textbf{Serialize}}(q^n.atoms)$
        \State $(q^n, l_g^n)\gets M_\phi(L^n)$
        \State $p_g^n\gets l_d\cdot l_g^n$
    \EndIf
\EndFor
\State $T\text{.append}((q^0, 1, 0))$ \Comment{directly use $q^0$ as root}
\State $T \gets \text{\textbf{ReIndexByBFS}}(T)$
\State \Return $T$
\end{algorithmic}
\label{algo:build-tree}
\end{algorithm}

\section{Reimplementation of TransferNet}
\label{app:transfernet}
To reimplemente TransferNet, we build the mixed relation graphs that consist of both label-form relations (i.e., KB triples) and text-form relations for KQA Pro and Musique, respectively, and train the models with the open source code. We show the details of graph building as follows.
\paragraph{KQA Pro}
We follow the method used by the original paper on MetaQA to build the relation graph of KQA Pro. As mentioned in Section~\ref{sec:implement}, we use half of its KB triples as the label form. We constructe the text form by extracting sentences from Wikipedia. Following the original paper, we use exact match of surface forms for entity recognition and linking. For every entity in the KB, we recall all the paragraphs in Wikipedia titled by it, then take the entity as subject and other relevant entities appeared in these paragraphs as objects. The sentences that contain the objects are selected as the relation texts. The recall of answer is 51\%, i.e, for 51\% questions, there exist a complete path from the topic entity to the answer in the relation graph, and this is a upper bound for the performance of TransferNet. 
\paragraph{Musique}
For each question in Musique, we utilize the 20 given paragraphs to build individual relation graph. Specifically, we first identify entities mentioned in these paragraphs via Spacy~\cite{honnibal2020spacy} and exact match of surface forms with Wikidata entities. Then we take the co-occuring sentences of two entities as the text-form, and take the triples in Wikidata whose subject or object is one of these entities as the label-form. The recall of answer is 72\%.